\documentclass[10pt, a4paper]{article}
\usepackage{lrec2022} 
\usepackage{multibib}
\newcites{languageresource}{Language Resources}
\usepackage{graphicx}
\usepackage{tabularx}
\usepackage{soul}
\usepackage{xcolor}
\usepackage{amsmath}
\usepackage{todonotes}
\usepackage{booktabs}
\usepackage{multicol}
\usepackage{comment}
\usepackage{titlesec}
\titleformat{\section}{\normalfont\large\bfseries\center}{\thesection.}{1em}{}
\titleformat{\subsection}{\normalfont\SmallTitleFont\bfseries\raggedright}{\thesubsection.}{1em}{}
\titleformat{\subsubsection}{\normalfont\normalsize\bfseries\raggedright}{\thesubsubsection.}{1em}{}
\renewcommand\thesection{\arabic{section}}
\renewcommand\thesubsection{\thesection.\arabic{subsection}}
\renewcommand\thesubsubsection{\thesubsection.\arabic{subsubsection}}

\newcommand{\norex}[1]{\textit{#1}}
\newcommand{\eng}[1]{`#1'}

\newcommand\N[1]{\textit{NorDiaChange}}

\usepackage{epstopdf}
\usepackage[utf8]{inputenc}

\usepackage{hyperref}
\usepackage{xstring}

\usepackage{color}

\title{NorDiaChange: Diachronic Semantic Change Dataset for Norwegian}

\name{Andrey Kutuzov\textsuperscript{1}, Samia Touileb\textsuperscript{2}, Petter Mæhlum\textsuperscript{1}, Tita Ranveig Enstad\textsuperscript{1}, Alexandra Wittemann\textsuperscript{1}} 

\address{\textsuperscript{1} University of Oslo, \textsuperscript{2} University of Bergen \\
         Norway \\
         andreku@ifi.uio.no, Samia.Touileb@uib.no, pettemae@ifi.uio.no, titare@ifi.uio.no, alexankw@ifi.uio.no\\
         }

\abstract{
We describe \N{}: the first diachronic semantic change dataset for Norwegian. \N{} comprises two novel subsets, covering about 80 Norwegian nouns manually annotated  with graded semantic change over time. Both datasets follow the same annotation procedure and can be used interchangeably as train and test splits for each other. \N{} covers the time periods related to pre- and post-war events, oil and gas discovery in Norway, and technological developments. The annotation was done using the DURel framework and two large historical Norwegian corpora. \N{} is published in full under a permissive licence, complete with raw annotation data and inferred diachronic word usage graphs (DWUGs).
\\ \newline  }

\begin{document}

\maketitleabstract


\section{Introduction} \label{sec:intro}
In this paper, we present \N{}: the first dataset of diachronic semantic change on the lexical level for Norwegian. Such datasets are required to evaluate lexical semantic change detection systems or contextualized embeddings in general, but can also be useful for historical linguists. \N{} is naturally accompanied with publicly available historical corpora.  

These historical corpora were used to produce \N{} via a meticulous manual annotation effort following the DWUG (Diachronic Word Usage Graphs) methodology \citelanguageresource{DWUG}: in short, it means that the annotators are yielding judgements about how semantically similar a word $x$ is in sentence pairs shown to them. As such, \N{} is fully compatible with datasets for other languages used, for example, in the SemEval'2020 shared task on semantic change detection \cite{schlechtweg-etal-2020-semeval}. However, it is different from most of them in that \N{} features two independent Norwegian datasets dubbed here `Subset 1' and `Subset 2'. Subset 1 deals with semantic change occurring between the period of 1929-1965 and the period of 1970-2013. Subset 2 focuses on the time periods of 1980-1990 and 2012-2019. Since the annotation procedure was exactly the same for both subsets, they can be used as train-test splits for each other.  

\N{} is published in full\footnote{\url{https://github.com/ltgoslo/nor_dia_change}} with all the raw annotation judgements so that any preferred scoring workflow can be applied to it. However, we stick to the standard DWUG scoring procedure described below and provide a graded and a binary change score to each target word.

The rest of the paper is structured  as follows. In section~\ref{sec:related}, we put our annotation effort in the context of prior work on semantic change detection. Section~\ref{sec:corpora} describes the historical corpora of Norwegian we used to sample sentences for annotation. In section~\ref{sec:target}, we explain how we selected the target words to annotate. Further on, section~\ref{sec:annotation} describes the annotation process itself. In section~\ref{sec:analysis}, we conduct qualitative linguistic analyses of the annotation results and sanity-check the output of the scoring algorithms we used. Finally, in section~\ref{sec:conclusion}, we conclude and discuss possible future avenues for our work.

\section{Related work} \label{sec:related}
Studying semantic change is a venerable field in linguistics; see \cite{bloomfield} and \cite{blank1999historical}, among many others.  However, in natural language processing, the topic of automatically tracing semantic change received comparatively little attention until the advent of easy-to-use distributional representations of lexical meaning (word embeddings). \newcite{kulkarni2015statistically}, \newcite{hamilton-etal-2016-cultural}, \newcite{hamilton-etal-2016-diachronic} and others have shown the potential that distributional semantics has for this task. We refer interested readers to comprehensive reviews on the topic: one can mention \cite{kutuzov-etal-2018-diachronic} and \cite{nina_tahmasebi_2021_5040302}, among others.

In order to compare different semantic change detection methods, one needs access to high-quality evaluation datasets. Although attempts to create such resources started as early as at least in 2011 \cite{baroni:2011}, they were diverse, not standardised, and suffered from sparse and biased data selection.

Currently, the mainstream approach to avoid this problems is to employ \textit{graded} contextual word meaning annotation, with the DURel framework \cite{schlechtweg-etal-2018-diachronic} being the most prominent example. In it, annotators are shown contextualized word usage pairs and asked to judge semantic similarity of two usages for the same word on a graded scale. After that, a change score is inferred from these judgements (see Section~\ref{sec:annotation} for more details). SemEval'2020 shared task on unsupervised semantic change detection employed datasets annotated within this framework, further developing it to include smart sampling of usage pairs and clustering word usages based on  their relations to each other in a usage graph \cite{schlechtweg-etal-2020-semeval}. 

A large trove of diachronic \textit{word usage graphs} annotated this way exists for several languages (English, German, Swedish, Latin) \cite{schlechtweg-etal-2021-dwug}, but this set lacks Norwegian. \N{} is aimed at filling in this gap, by being fully compatible to the existing datasets (produced using the same procedure). 

Additionally, we are following \cite{rodina-kutuzov-2020-rusemshift} and \cite{kutuzov-pivovarova-2021-three} in providing \textit{several} independent datasets for a particular language (in our case, Norwegian). These \textit{subsets} are different in their target time periods and word lists, but are annotated in the same way and by the same team. This potentially allows to develop \textit{supervised} semantic change detection systems which can be trained or fine-tuned on one subset and then evaluated on another. Systems using fine-tuning instead of  `zero-shot' approaches turned out to be the best in the recent semantic change detection shared task for Russian \cite{rushifteval2021}, and so we consider this possibility to be crucial for our \N{} dataset as well.

\section{Corpora used} \label{sec:corpora}

The underlying corpora of our work are the NBdigital\footnote{\url{https://www.nb.no/sprakbanken/ressurskatalog/oai-nb-no-sbr-34/}} corpus from the National Library of Norway, and the Norwegian newspaper corpus (Norsk Aviskorpus or NAK\footnote{\url{https://www.nb.no/sprakbanken/ressurskatalog/oai-nb-no-sbr-4/}}). Both corpora are freely available, and can be downloaded from the website of the  National Library of Norway. The NBdigital is a historical corpus containing a collection of over 26,000 books, reports, and news articles from the public domain. The texts cover various genres, written in various languages including both Norwegian forms (Bokmål and Nynorsk). In addition, each text has a list of metadata information, as e.g. author, OCR confidence, and date of publication. However, due to the use of OCR, not all of the texts are of acceptable quality. We therefore had to first filter out all texts with an OCR confidence below 70\%, and thereafter cleaned the non Norwegian Bokmål texts from the Bokmål collection of NBdigital. 

The content of NBdigital is dated up to 2013. In order to get more recent data that can reflect more modern language use, we have also selected news articles from the NAK corpus. We therefore only use articles published between 2012 and 2019 from NAK.

We decided to divide the corpus into two subsets. The first subset compares the time periods of 1929-1965 and 1970-2013. In the second subset, we look at the time period of 1980-1990 compared to 2012-2013. In what follows, we give the arguments behind this decision, and describe the content of each subset. 

\subsection{Subset 1: 1929-1965 VS 1970-2013}

This subset captures two important historical time periods for Norway. The pre- and post-war periods have affected Norwegians' standard of living, and therefore their language use. Higher living standards and better economy after the 1960s made more people, traditionally farmers, move to bigger cities\footnote{\label{ssb}Statistics Norway: \url{https://www.ssb.no/en/befolkning/artikler-og-publikasjoner/_attachment/364602?_ts=1664418b978}}. By the advent of the 1970s a consumer culture was established and new technology entered Norwegian homes. All of these changes impacted the language use by both introducing new words to the vocabulary, and adding new senses to pre-existing words. 
The time period between the 1970s and 2013 also introduced quite many technologically related words and senses, and as society developed, the language arguably followed the trends.

In this subset, both time periods are extracted from the historical NBdigital corpus. Table \ref{tab:subset1} shows the total number of words and documents in both time periods of Subset 1.

\begin{table}
{   \centering
    \begin{tabular}{lrr}
    \toprule
        Period & Words & Documents \\
         \midrule
         1929 -- 1965 & 57 mln & 959 \\
         1970 -- 2013 & 175 mln & 4,209 \\
        \bottomrule
    \end{tabular}
    \caption{Total number of word tokens and documents in Subset 1.}
    \label{tab:subset1}}
\end{table}

\begin{table}
{   \centering
    \begin{tabular}{lrr}
    \toprule
        Period & Words & Documents \\
         \midrule
         1980 -- 1990  & 43 mln & 1,115 \\
         2012 -- 2019 & 649 mln & 1,763,843  \\
        \bottomrule
    \end{tabular}
    \caption{Total number of word tokens and documents in Subset 2.}
    \label{tab:subset2}}
\end{table}

\subsection{Subset2: 1980-1990 VS 2012-2019}

This subset contains shorter time periods than in Subset 1, but we still expect shifts in word usages. The changes between the two periods of 1980-1900 and 2012-2019 can be caused both by linguistic and cultural factors. Most of the changes are expected to be within technology, and the changes in language use are certainly mostly in vocabulary additions. However, many words related to technological advances were added as new senses to pre-existing Norwegian words. 

The first time period is extracted from NBdigital, while the second one from 2012 to 2019 comprises texts from the NAK corpus. The language use in NAK might be different from NBdigital, as we expect news texts to contain more `modern' senses of words that have shifted from the previous time period. 
In Table \ref{tab:subset2}, we give an overview of the total number of words and documents in each time period of Subset 2.

\section{Choosing target words} \label{sec:target}

We manually selected target words that we believe can have undergone semantic changes during the periods of Subset 1 and Subset 2. This was based on linguistic intuition of the authors as native Norwegian speakers and on existing linguistic work, similar to what has been done by \newcite{rodina-kutuzov-2020-rusemshift} and \newcite{schlechtweg-etal-2021-dwug}. Another option would be to look at diachronic dictionaries, but \newcite{VikoerLarsS2006Ddav} notes that although some dictionaries make note of antiquated senses for some lemmas, there are no truly diachronic dictionaries for Norwegian, with the exception of etymological dictionaries. 

For each of the selected target words, a filler word was added to the word list. Filler words were randomly sampled from the corresponding corpora; their frequency percentiles in both earlier and later time periods had to be similar to those of the corresponding target words (in this, we followed the current best practices). The purpose was to ensure that corpus frequency dynamics alone cannot be used to solve the dataset (frequency changes often accompany semantic changes). 

Indeed, there are no statistically significant correlations between frequency differences (IPM-normalised) and annotated change scores. For Subset 1, Spearman $\rho$ is $-0.06$ ($p=0.70$), for Subset 2 it is $0.11$ ($p=0.49$). It means that \N{} does not just list words which sharply changed their corpus frequency together with semantic shifts. It is balanced with regards to word frequencies, and the systems aiming to approximate the scores in \N{} based on corpus data must take into account more than that.

Note that the filler words were manually checked to make sure that they are valid Norwegian lemmas not immediately associated with any known diachronic semantic shift. However, after the annotation, we found out that some of the filler words actually did experience some historical change; see section~\ref{sec:analysis} below.
All our target and filler words are nouns and we discarded words belonging to other parts of speech. 

The initial target word lists for both subsets (without the fillers) contained 20 words each (40 target words in total). These words are expected to have semantically shifted between 1929-1965 and 1970-2013 for Subset 1, and between 1980-1990 and 2012-2019 for Subset 2. For some of the words, we also did a dictionary check to see if the word seemed to have lost senses in different time periods. 
The full list (including senses) can be found in the project GitHub repository.

\section{Annotation process} \label{sec:annotation}

The data was annotated using the DURel framework \cite{schlechtweg-etal-2018-diachronic} and the accompanying web service.\footnote{\url{https://www.ims.uni-stuttgart.de/data/durel-tool}} The annotators were presented with \textit{word usage pairs}: two snippets of text, both containing the same target word. The annotation task is to decide the semantic similarity of the target word in the two different contexts, using the following graded scale:
\begin{itemize}
    \item senses are \textit{identical}: the word usage pair is annotated with $4$
    \item senses are \textit{closely related}: $3$
    \item senses are \textit{distantly related}: $2$
    \item senses are \textit{unrelated}: $1$
    \item the annotator is unsure and cannot decide the semantic similarity: $0$
\end{itemize}

Thus, annotators are judging usage pairs on a semantic proximity scale, avoiding any manual definitions of word senses. An example of semantically unrelated usages is the word \norex{ris} in the following usage pair:
\begin{enumerate}
    \item \norex{Sidene pagineres i hver bok fra 1—500: Papiret koster pr. \textbf{ris} å 500 ark kr. 32,80, hvorpå beregnes 20 pet. fortjeneste.} \\
        (\eng{The pages are numbered from 1-500 in each book: The paper costs for each \textbf{ream} of 500 sheets NOK 32,80, to which a 20\% profit is calculated.})
    \item \norex{I saltvann produseres primært grønn tang til en samlet verdi av \$ 440 mill. (1973). Denne tangen spises til \textbf{ris} som en slags grønnsak.} \\
    (\eng{In seawater green seaweed is primarily produced for a total value of \$ 440 million (1973). This seaweed is eaten with \textbf{rice} as a type of vegetable.})
\end{enumerate}

In the first sentence, \norex{ris} refers to a unit of quantity of paper, while in the second sentence, it refers to rice. The annotator is expected to yield the $1$ judgement.

Contextualised word usages are essentially sentences containing at least one target word (can be a filler). These sentences were sampled from the historical corpora described in section~\ref{sec:corpora} in the following way. First, we lemmatized and POS-tagged our corpora using UDPipe \cite{straka-strakova-2017-tokenizing}. This was required to be able to find inflected forms of the target words. From the processed corpora, we extracted all sentences containing at least one token with a target word as its lemma (we filtered out target words not tagged as a noun).  The extracted sentences (both raw text and processed versions with lemmas and POS tags) were stored as time period marked JSON files for further usage: one file per target word / time period.

From each of these files, we randomly sampled $11$ sentences as representatives of a particular target word usage in a particular time period. Before sampling, the sentences were de-duplicated, and we discarded sentences containing the `\^{}` character (in most cases, it signalled heavy OCR errors). This means that for every target word, we had $22$ total usage examples ($11$ from the earlier and $11$ from the later time period). In the rare cases when there were less than $11$ occurrences of the target word\footnote{\norex{Idiot} and \norex{katten} in Subset 1; \norex{fane} and \norex{syden} in Subset 2. It has never been less than $7$ occurrences per a time period.}, all the existing occurrences were used.

$11$ might seem to be a small sample: \newcite{schlechtweg-etal-2021-dwug} sampled 100 usages for each target word / historical corpus. We are aware of the limitations of our sample size, and that we cannot guarantee that all senses are covered. However, one should keep in mind that not all possible usage pairs are getting annotated anyway (see more on that below). In the end, \N{} contains $26,730$ annotator judgements, which is comparable to the existing datasets for English ($29,000$) and Swedish ($20,000$).

\begin{figure}
    \centering
    \includegraphics[width=0.8\linewidth]{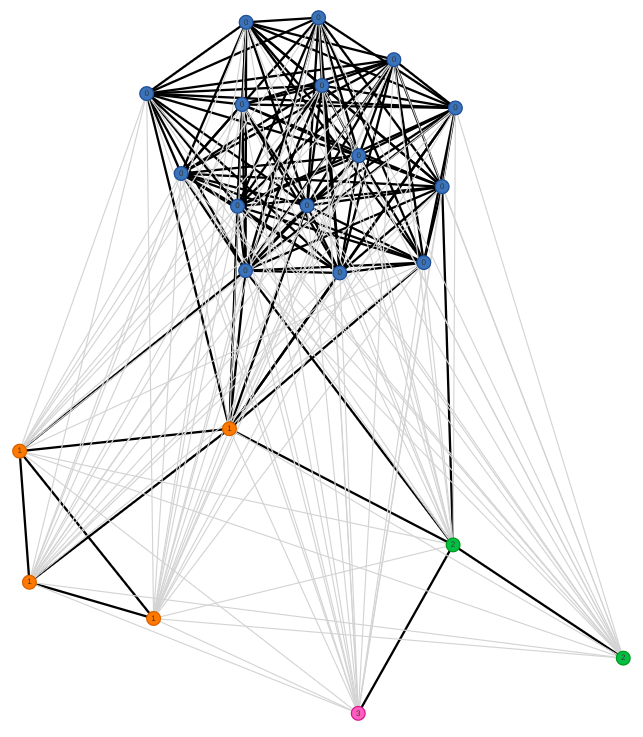}
    \caption{Word usage graph for \norex{innstilling}, time periods 1980-1990 and 2012-2019.}
    \label{fig:instilling}
\end{figure}

The sampled sentences were fed to the DURel web service which was responsible for randomly selecting usage pairs to present to the annotators. More details on that can be found in \cite{schlechtweg-etal-2021-dwug}. In short, the aim here was to spend as little annotation effort as possible to construct a well-connected diachronic word usage graph (DWUG), where nodes are usages (sentences) and edges between them are weighted with annotator's judgements. Thus, sentences where a target word is used in roughly the same sense ($4$ judgements), are naturally grouped into a `sense cluster'. See, for example, DWUG for the Norwegian word \norex{instilling} in figure~\ref{fig:instilling}. Node colours correspond to automatically inferred sense clusters, edge thickness corresponds to the median annotators' judgements. The blue cluster contains sentences where \norex{instilling} is used in the older sense of `recommendation', and they come from both 1980-1990 (earlier) and the 2012-2019 (later) time periods. However, the orange, green, and lilac clusters contain exclusively sentences from the later time period, with \norex{instilling} used in the senses of `setting' (as in `account settings') or `attitude'. Thus, this DWUG constitutes a case of a word gaining new senses diachronically.

\N{} is published with full raw annotation data, so it is possible to infer sense grouping from human similarity judgements on usage pairs in any preferred way. However, we followed the standard workflow from \cite{schlechtweg-etal-2021-dwug} and employed their code to produce clustered DWUGs. After the sense clusters are inferred, it becomes possible to compare sense distributions across time (normalised to become probability distributions). The graded change score was the main score we assigned each word in \N{}. It is calculated as Jensen-Shannon distance (JSD) between the probability distributions of senses in the earlier and the later time periods \cite{schlechtweg-etal-2019-wind,giulianelli-etal-2020-analysing}. 

To continue the example with \norex{instilling} on figure~\ref{fig:instilling}, there are four sense clusters in total. In the earlier time period, all $11$ usages belong to the first cluster, so the usage distribution is $[11, 0, 0, 0]$ or $[1, 0, 0, 0]$ when converted to a probability distribution. In the later time period, the usage distribution is $[4, 4, 2, 1]$ (only $4$ of the usages belong to the first cluster), and after converting to a probability distribution, it is $[0.364, 0.364, 0.182, 0.091]$. The Jensen-Shannon distance between $[1, 0, 0, 0]$ and $[0.364, 0.364, 0.182, 0.091]$ equals to $0.655$ which is the degree of diachronic change that the word \norex{instilling} experienced between 1980-1990 and 2012-2019.

Note that due to essentially random sampling of sentence pairs, some potential groupings of nodes (sentences) might end up unconnected with edges. This can in theory negatively influence the clustering. This is why we also calculated the number of uncompared cluster combinations for each target word. It is zero in almost all the cases, with the exception of one word in the Subset 1 (\norex{bølge}) and two words in the Subset 2 (\norex{kanal} and \norex{sky}).

In addition, we provide binary change scores, where each word is assigned a $1$ label if it gained or lost any senses between two time periods or a $0$ label if its senses remained stable. In determining these scores, we again followed the approach from \cite{schlechtweg-etal-2020-semeval}. A word is considered to experience a binary change, if at least one sense was attested at most $k$ times in one time period and at least $n$ times in another time period. $n$ and $k$ here are manually defined hyper-parameters which are needed to filter out insignificant fluctuations in sense frequencies. We kept the default values of $k=1$ and $n=3$. This means that, for example, having an entirely novel sense cluster with two usages in it is not enough to assign a $1$ binary change score (since $2 < 3$). However, in the case of \norex{instilling}, \N{} marks it as binary changed, since its second sense is represented with $4$ usages in the later time period ($4 \ge 3$) and $0$ usages in the earlier time period ($0 \le 1$).

\subsection{Annotators}

\N{} was annotated by three native speakers of Norwegian, all of whom hold bachelor degrees in either linguistics or language technology. They received identical guidelines which can be found at \url{https://github.com/ltgoslo/nor_dia_change/blob/main/guidelines.md}. Several reconciliation meetings were held between the annotators and project managers to make sure that all annotators understand the guidelines in the same way, and to clear up inconsistencies in the data. Each annotator produced from $9,167$ to $10,584$ judgements in total and was paid about $13,000$ Norwegian Krone (about $1,300$ euros) for their job after taxes.

\subsection{Annotation results}

Each usage pair received $1.8$ annotator's judgements on average, so possible errors or misunderstandings of one annotator could be compensated by another. The most important descriptive statistics for both \N{} subsets are given in Table~\ref{tab:decriptive_stats}; more is available at our repository. For comparison, we also provide the same statistics for the Swedish dataset from \cite{schlechtweg-etal-2021-dwug} and the Russian \textit{RuShiftEval-2} dataset from \cite{kutuzov-pivovarova-2021-three}. 

\begin{table}
{   \centering
    \begin{tabular}{lccccc}
    \toprule
         \textbf{Dataset} & \textbf{Words} & \textbf{$\lvert U \rvert$} & \textbf{JUD} & \textbf{SPR} & \textbf{KRI} \\
         \midrule
            \textbf{Subset 1} & 40  & 21 & 12,727 & 0.77 & 0.76 \\
            \textbf{Subset 2} & 40  & 21 & 14,003 & 0.71 & 0.67 \\
        \midrule
        \textbf{Swedish} & 40 & 168 & 20,000 & 0.57 & 0.56 \\
        \textbf{Russian} & 99 & 60 & 8,879 & 0.56 & 0.55 \\
        \bottomrule
    \end{tabular}
    \caption{Descriptive statistics for the full \N{} subsets. $\lvert U \rvert$ is the average number of usages sampled for a word from the corpora. JUD is the total number of annotators' judgements for a subset. SPR is the weighted mean of pairwise Spearman $\rho$ correlations between different annotators, and KRI is the value of Krippendorff’s $\alpha$ inter-rater agreement.} 
    \label{tab:decriptive_stats}}
\end{table}

Table~\ref{tab:top} shows top $5$ most changed words for both \N{} subsets. The `graded' column gives the values of the graded change score (calculated with JSD). The `sense gain' and `sense lost' columns demonstrate a binary label depending on whether a word acquired or lost at least one sense based on the binary change score calculated with the same default values of $k=1$ and $n=3$.  

It is interesting to note that one of those words in Subset 1 (\norex{horisont}) and three in Subset 2 (\norex{stryk}, \norex{oppvarming} and \norex{innstilling}) were filler words, randomly sampled from the corpora: that is, we did not intentionally came up with these words as `changed'. This is normal (the same was observed, for example, by \newcite{kutuzov-pivovarova-2021-three}) and demonstrates that mining real textual data can sometimes yield unexpected but still valid insights. However, in general human annotations correspond well to our original intuitions: if we assign the value of $1.0$ to the words originally chosen as target ones and $0.0$ to the words originally sampled as fillers, then both subsets show highly statistically significant ranked correlations of these values and the graded change scores. For Subset 1, Spearman $\rho$ in this case is $0.38$ ($p=0.01$) and for Subset 2, it is $0.40$ ($p=0.02$). 

\begin{table}
{   \centering
    \begin{tabular}{lccc}
        \toprule
         \textbf{Word} & \textbf{Graded} & \textbf{Sense gain} & \textbf{Sense lost}  \\
         \midrule
         \multicolumn{4}{c}{\textbf{Subset 1} (1929-1965 VS 1970-2013)} \\
         \midrule
         \norex{plattform} & 0.87 & 1 & 1\\
         \norex{leilighet} & 0.80 & 0 & 1\\
         \norex{horisont} & 0.64 & 1 & 1\\
         \norex{mål} & 0.60 & 0 & 1\\
         \norex{bølge} & 0.60 & 1 & 0\\
         \midrule
         \multicolumn{4}{c}{\textbf{Subset 2} (1980-1990 VS 2012-2019)} \\
         \midrule
         \norex{stryk} & 1.00 & 1 & 1 \\
         \norex{kanal} & 0.73 &  0 & 1 \\
         \norex{kode} & 0.73 & 1 & 1\\
         \norex{oppvarming} & 0.72 & 1 & 0 \\
         \norex{innstiling} & 0.66 & 1 & 0\\
         \bottomrule
    \end{tabular}
    \caption{Top changed words in \N{}.} 
    \label{tab:top}}
\end{table}

As it was expected based on prior work, some target words received disproportionately many $0$ annotations, for different reasons.\footnote{Overall, there were $548$ zero judgements in Subset 1 and $286$ in Subset 2 (Subset 1 suffers much heavier from OCR errors).} In some of those cases, a target word has a verbal homonym which was erroneously tagged as a noun by UDPipe. All occurrences where the target word was a verb in a word usage pair were annotated with 0. For the target word \norex{vert} \eng{host} from Subset 2 , 27\% of the examples were actually the verb \norex{verte} \eng{become}. This resulted in the word usage graph for 1980-1990 only containing three nodes (usages with high proportion of zero judgements are removed from the graph automatically). The annotation of the word \norex{tap} \eng{loss} from Subset 2 also yielded small word usage graphs due to many $0$ annotations. This was in part because 9\% of the context examples were actually the verb \norex{tape} \eng{lose}.  

On the other hand, the word \norex{fil} \eng{file} from Subset 1 had many $0$ annotations due to noisy data. There were so many examples where it was impossible to interpret the context, that the word usage graph for 1929-1965 only had three nodes, all of them in the same cluster.

\begin{figure*}[!ht]
    \centering
    \includegraphics[width=0.7\textwidth]{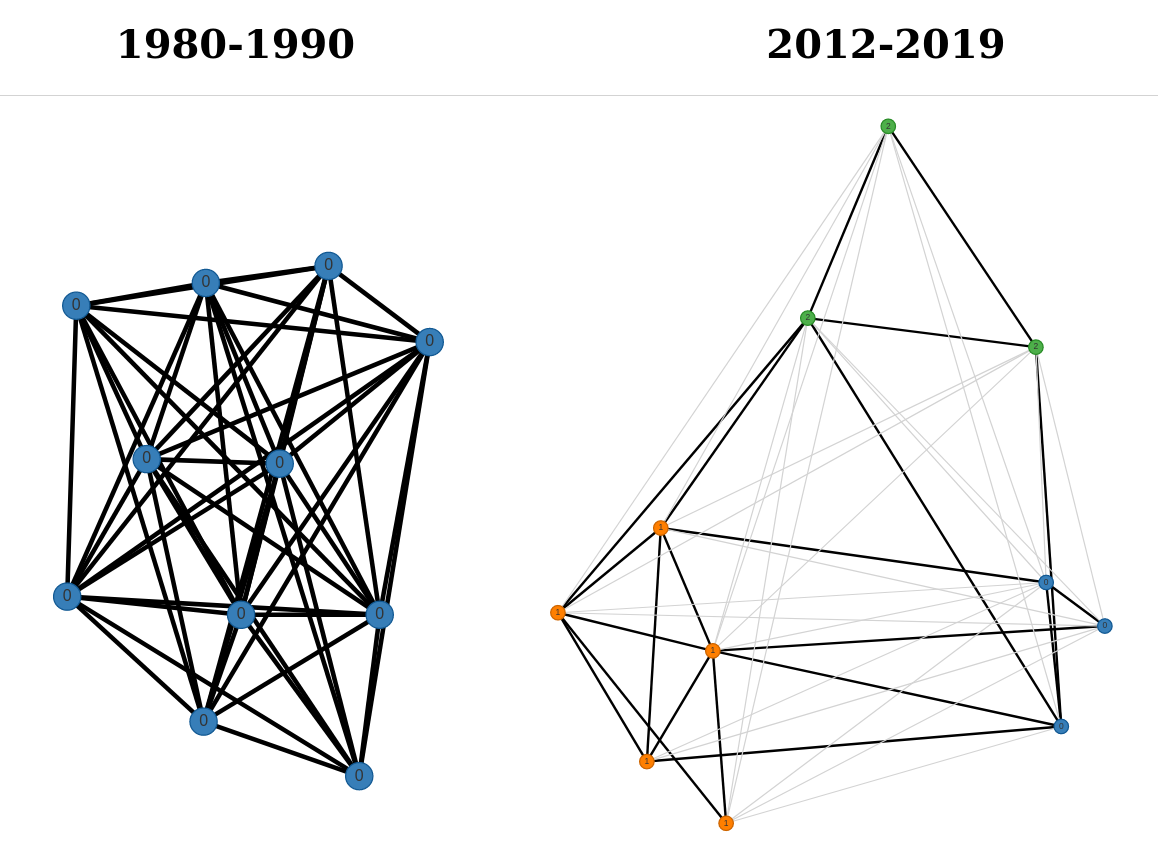}
    \caption{The two word usage graphs for \norex{oppvarming} aligned. The left graph shows the earlier time period (1980-1990), with the only sense of \eng{heating} (0). The right graph shows later time period (2012-2019), with the three clusters referring to \eng{heating} (0), \eng{global heating} (1) and \eng{warm-up} (2).}
    \label{fig:aligned}
\end{figure*}

We also had some words where the team didn't fully agree with the produced results. The usage graph for the word \norex{mening} \eng{meaning, opinion} from Subset 1 clustered together senses that should not have been in the same cluster. As a result, the scoring algorithm also marked this word as binary changed, contrary to the opinion of the annotators. During a reconciliation meeting, the team agreed that the nuances in the word semantics were difficult to annotate due to the abstract nature of the word. 

The annotation of the word \norex{damp} \eng{steam} from Subset 1 yielded a somewhat `strange' word usage graph for the time period 1929-1965. The graph contains four clusters, but two of these have the same sense (steamboat) and should have been grouped together into one cluster. 

For the word \norex{mus} \eng{mouse} from Subset 2, the annotation showed that the sense meaning computer mouse had disappeared, but all native speakers agree that this sense is still very much present in the language. See section \ref{sec:analysis} for more details. We decided to mark these problematic words as questionable, and publish both a `clean' dataset with 37 target words in each subset, and the full set with all 40 annotated words in each subset, including the six `questionable' ones.

\section{Meaning change in the Norwegian language of the 20th century} \label{sec:analysis}

In order to qualitatively analyse the binary and graded change scores produced within \N{}, we consulted the annotators and authors whose native language was Norwegian. Their intuitions are the basis of these evaluations. The annotators consulted the graphs for each period, as in Figure~\ref{fig:aligned}, and looked at the sense clusters to examine which senses had been clustered in which ways. However, it can be difficult, even for native speakers, to personally judge the semantic graphs. In an attempt to mitigate native speaker bias,  we believe that looking at frequencies over time can be one way to get additional insight into the DWUGs. The National Library of Norway has released the National Library N-gram (NB N-gram) service, which allows users to check the corpus frequencies of an n-gram between 1810 and 2013. The data is similar to what was used for this annotation effort, with the exception of the NAK-data for the latter part of subset 2.\footnote{\url{https://www.nb.no/sprakbanken/ressurskatalog/oai-nb-no-sbr-70/}}. We use this service to highlight certain insights. The following discussion is based on the \eng{clean} dataset.

\begin{figure}
    \centering
    \includegraphics[width=0.4\textwidth]{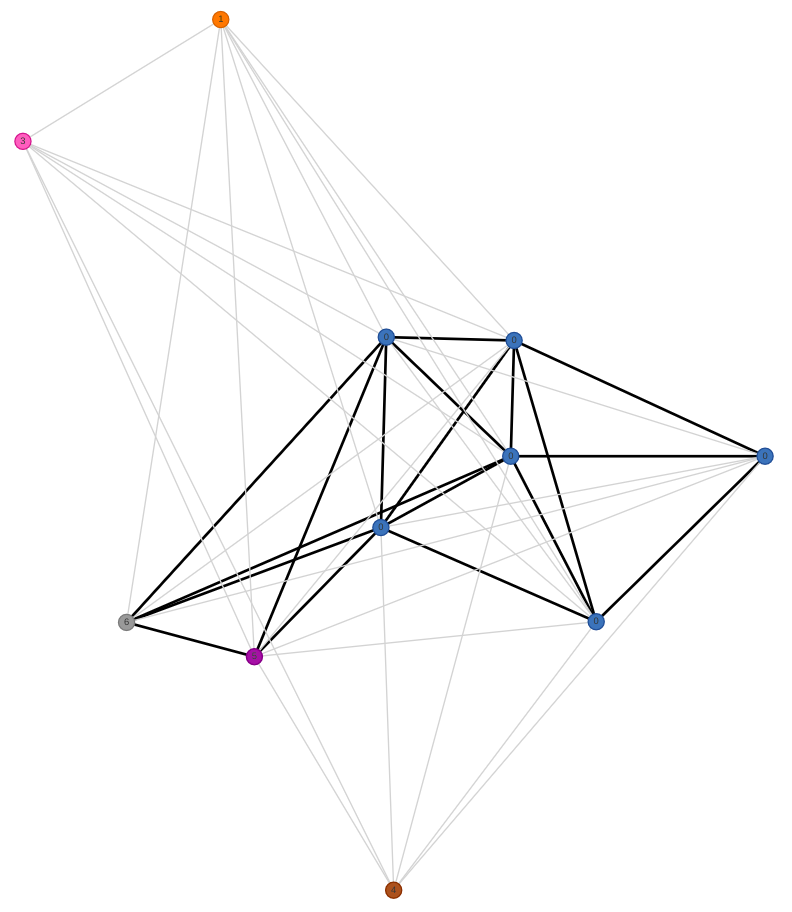}
    \caption{Sense clusters for \norex{bølge} \eng{wave} from the first time period (1929-1965) The blue cluster marked 0 represents the \eng{(sea) wave} sense, the remaining senses represent metaphorical usage. }
    \label{fig:boelge}
\end{figure}

The annotators all agree on the binary results, but there are parts of the sense graphs that do not necessarily fit the annotators' intuition completely. One case is when a sense is overly specific. An example of this can be seen for the word \norex{oppvarming} \eng{heating} (figure~\ref{fig:aligned}), where in the later period of 2012-2019, the new sense \eng{(global) warming} contributes to about 5/8 of the graded change score. However, this sense is closely related to the \eng{heating} sense, and might not be seen as separate. 
It seems like the graph is sensitive to what might be perceived as different types of metaphorical usage. This is not necessarily an error, but it is worth noting that this can potentially expand the numbers of senses considerably. Another example of this is \norex{bølge}, (see Figure~\ref{fig:boelge}) whose graph has as many as 6 senses. 5 of these are metaphorical: waves of gratitude, waves of battle, waves of hair, waves of grain, waves of song.   

The other case is when two senses are conflated. This can happen both within the sense graph for one period, or across time periods for senses that are considered the same.  One example is the graph for \norex{skjerm}, in which it seems like at least one of the examples for the \eng{computer screen} sense, based on its sense in the later time period, also contains a sense of \eng{canvas used when taking a photo}, or something related. This could also be due to difficulties during annotation, where annotators report that some senses are more difficult to judge from context than others. On the other hand, we used the default settings for graph clustering; tuning them may yield other groupings, more or less specific.

Another problem is frequency of senses in both corpora. Although the annotators agreed that it seems likely that the frequency of the sense \eng{rapids} have diminished greatly for \norex{stryk} in 2012-2019, they note that the sense is still in use, and that the largest actual change in the sense of \norex{stryk} is likely the addition of the \eng{fail} sense, rather than the absence of the \eng{rapids} sense. Overall, there seems to be a tendency for certain scientific senses to be more frequent in the earlier time periods. Examples of this is the \eng{geological horizon} sense for \norex{horisont}, the botanical sense of \eng{umbrella} for \norex{skjerm}, and perhaps \eng{distribution board} for \norex{tavle}. 

A common problem for semantic change is the tendency in Norwegian, as in Swedish and German, among other languages, to spell compound nouns as single tokens, without spaces. However, this makes exact matches with compounds containing these words impossible. In English, one might be able to match the word \norex{computer} in the compound \norex{computer game}, while this would not be possible in Norwegian. For example, according to Leksikografisk Bokmålskorpus (LBK)\footnote{\url{https://www.hf.uio.no/iln/om/organisasjon/tekstlab/prosjekter/lbk/}}  \citelanguageresource{bokmalsk},  the word \norex{kode} \eng{code} occurs word-initially in around 132 lemmas, and word-finally in about 138 lemmas. In some cases the changes associated with a lemma might be more or less visible in only certain compounds, even if the lemma itself has not lost its sense. It could possibly also be that a compound lemma is preferred to the more ambiguous lemma by itself. We do not claim any direct connection between the sense of a noun and the frequencies of its compounds, but believe they can be an indicator of usage. Some examples of this are discussed below.

\subsection{Subset 1}
Of the 37 words in (`clean') Subset 1, 11 words showed  binary semantic change. Details are shown in table~\ref{tab:subset_summary}. This period had 3 more words with binary change than Subset 2, resulting in a higher percentage of binary change. The average graded change is also higher. The words with binary change were \norex{anfektelse} \eng{distraction},
\norex{bit} \eng{bite},
\norex{bølge} \eng{wave},
\norex{forhold} \eng{relation, relationship},
\norex{horisont} \eng{horizon},
\norex{leilighet} \eng{opportunity, apartment},
\norex{mål} \eng{goal, measure},
\norex{pære} \eng{pear, bulb},
\norex{plattform} \eng{platform},
\norex{rev} \eng{fox},
\norex{skjerm} \eng{screen}. An additional 10 words had graded change score above zero. We will discuss three interesting cases from this period below.

\paragraph{Plattform} This word had three senses in the earlier time periods, and two in the latter. In the first period, the sense of \eng{generic platform} dominated with 5 cases, whereas 4 cases were of \eng{tram platform} and 1 was \norex{oil platform}. Not surprisingly, it is the \eng{oil platform} sense that dominates in the latter period. It is also interesting to see the disappearance of the \eng{tram platform} sense, which is likely due to changes in how the tram worked. This change seems typical of the period. 
 
\paragraph{Rev} Intended for its possible sense of \eng{joint}, the word \norex{rev} turned out to show other changes over time. Both time periods show frequent usage of the expected sense \eng{fox}, but we observe that the sense of \eng{reef} has become more frequent.

\paragraph{Pære} The sense of \eng{light bulb} seems to have been present in both periods, but with only 2 cases in the earlier period, and 9 in the latter. Interestingly, the \eng{pear (fruit)} sense becomes less frequent, according to our data. This is contrary to our original expectations, as one would expect there to be more mention of the electric bulbs. According to the N-gram service, the less ambiguous \norex{lyspære} \eng{light bulb} is present in both time periods, but steadily increasing, with a higher frequency in the latter period.

\begin{table}
{   \centering
    \begin{tabular}{lcccc}
    \toprule
         \textbf{Subset} & \textbf{Words} & \textbf{Binary} & \textbf{Percent.} & \textbf{Average} \\
         \midrule
            \textbf{Subset 1} & 37 & 11 & 29.7 & 0.26 \\
            \textbf{Subset 2} & 37 & 9 & 24.3 & 0.22 \\
        \bottomrule
    \end{tabular}
    \caption{Frequencies for subset 1 and subset 2, indicating the number of binary changes, the corresponding percentage of changed words, and the average graded change.
    }
    \label{tab:subset_summary}}
\end{table}

\subsection{Subset 2}

Of the 37 words in (`clean') Subset 2, 8 words showed binary semantic change, while 9 more showed graded change above zero. The remaining 23 words showed no change.
The nine words with binary change were \norex{stryk} \eng{rapids, fail}, \norex{kanal} \eng{channel}, \norex{kode} \eng{code}, \norex{oppvarming} \eng{heating}, \norex{innstilling} \eng{setting}, \norex{tavle} \eng{(black)board}, \norex{fane} \eng{banner}, \norex{strøm} \eng{current, electricity}. As with subset 1, we will look at four interesting words from this subset, including the `questionable' word \norex{mus} \eng{mouse}.

\paragraph{Stryk} was perhaps somewhat surprising. The only word with a perfect graded change score of 1, its score comes from the disappearance of the frequent sense \eng{rapids}, and the introduction of the \eng{fail} sense. The annotators note that the \eng{rapids} sense still exists, but the \eng{fail} sense is new. The national library N-gram service (NB N-gram) shows that indeed, the relative frequency of the word \norex{strykkarakter} \eng{fail grade}, chosen for its unambiguity while being etymologically related to \norex{stryk}, went drastically up towards the end of the 1990's, keeping a higher frequency after its peak.

\paragraph{Kanal} This word showed a high degree of change. In the earlier time period, it had 5 senses, where one roughly meaning \eng{TV and radio channel} became much more frequent in the later time period, while one new sense, seemingly something like \eng{communication channel}, appeared. All other senses disappeared from the word usage graph. Although it is unlikely that the other senses, such as \eng{river channel} and \eng{electric channel} have indeed disappeared from language in general, the increase in frequency for the \eng{TV and radio channel} sense unsurprisingly fits recent trends well.

\begin{figure}[ht]
    \centering
    \includegraphics[width=0.4\textwidth]{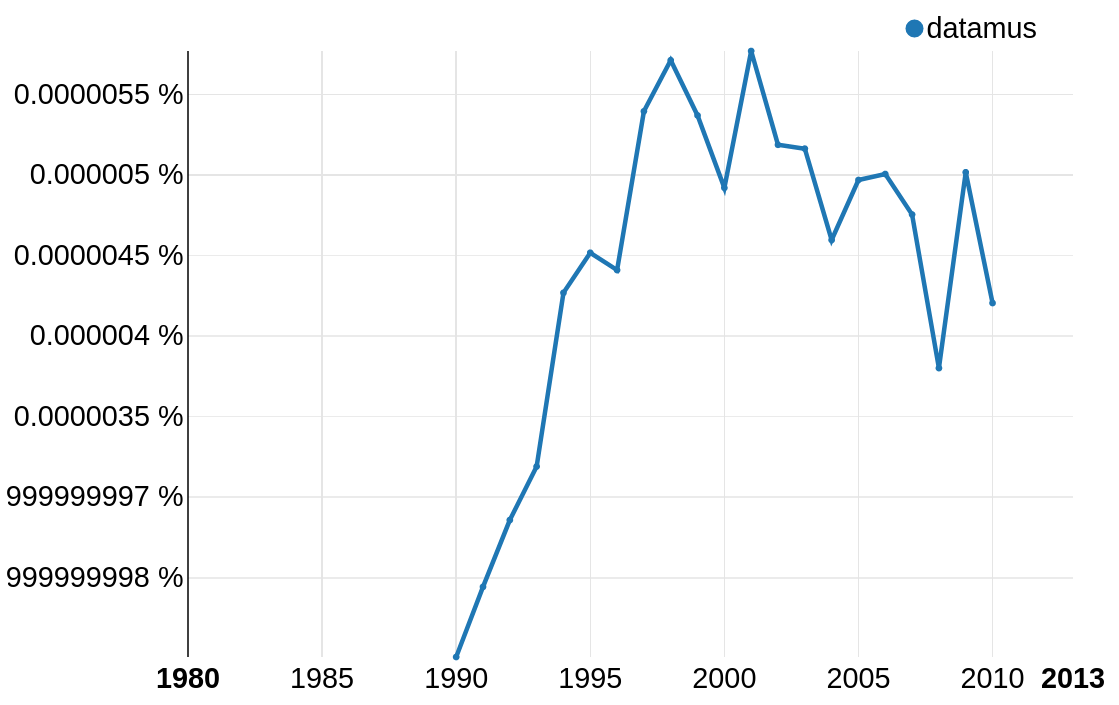}
    \caption{An example of a NB N-gram graph, in this case for the word \norex{datamus} \eng{computer mouse}. The graph shows that the word appeared in the corpus in 1990 and sharply increased its frequency.}
    \label{fig:datamus}
\end{figure}

\paragraph{Mus} Interestingly, this word lost the sense of \eng{computer mouse} in the later period, according to our usage graphs. This is the reason why it was judged as \eng{questionable}. The only sense present in the later time period is that of the animal. However, if we look at the compound \norex{datamus} \eng{computer mouse}, we see that the sense appears in 1990 and becomes more frequent (see figure~\ref{fig:datamus}). This strengthens the annotators belief that the loss of the \eng{computer mouse} sense might be due to data sampling error.

\paragraph{Spill}
Somewhat surprisingly, this word did only show a small degree of graded change. Both time periods are dominated by the general sense \eng{game}, and its change is mostly due to the loss of earlier senses. However, if we look at the frequency of the compound noun \norex{dataspill} \eng{computer game} in the NB N-gram service, we see that although there are a few instances between 1960 and 1980, it is not until the late 1980's that the word becomes more frequent, and it is much more frequent after the 1990's than before. 
\paragraph{Fane} \eng{banner, plume} is a good example of how the recent year's digital changes have affected word senses. The word has two senses in the earlier period: \eng{banner} and \eng{plume of smoke}. In the newer period, the \eng{banner} sense prevails, although it seems like its use is somewhat more metaphorical, and in addition we can see the new sense \eng{(website) banner}. 

\section{Conclusion} \label{sec:conclusion}
We introduced \N{}, a dataset of diachronic semantic change on the lexical level for Norwegian. The dataset comprises two new manually annotated subsets that can be used interchangeably as train or test splits. We followed the mainstream DURel framework during annotation, using two historical Norwegian corpora. All the data in the form of diachronic word usage graphs (DWUGs) and accompanying code is available under a CC-BY license at \url{https://github.com/ltgoslo/nor_dia_change/}. We conducted comprehensive qualitative analysis of the annotation results.

This is the first attempt at developing resources for Norwegian diachronic semantic change, and \N{} will become one of the standard NLP benchmarks for Norwegian.
In the nearest future, we plan to evaluate various semantic change modelling systems on \N{}, and report baseline performances. We are also going to further assess available Norwegian historical text collection in order to come up with a set of reference corpora which are more comparable in terms of their size and genre distribution than the ones we used for \N{}.

It is important to note that \N{} can be used not only in the field of lexical semantic change detection. By definition, it is also a full-fledged WiC (`word-in-context') dataset \cite{pilehvar-camacho-collados-2019-wic}. As such, it is a ready-to-use benchmark to evaluate \textit{synchronic} word sense disambiguation capabilities of pre-trained language models for Norwegian. Simultaneously, the same models can in theory be used to automatically `annotate' the existing data, yielding DWUGs for many thousands of words, not just dozens. This is another direction for our future research.
Finally, we plan to extend the dataset beyond nouns.

\section*{Acknowledgements}
The annotation of \N{} was kindly funded by the Teksthub initiative at the University of Oslo.
We thank the three annotators Helle Bollingmo, Tita Ranveig Enstad, and Alexandra Wittemann for all their hard work and contributions. A special thanks to Ellisiv Gulnes Heien who provided us with some of the target words. 

\section{Bibliographical References}\label{reference}

\bibliographystyle{lrec2022-bib}
\bibliography{anthology,lrec_lscd}

\section{Language Resource References}
\label{lr:ref}
\bibliographystylelanguageresource{lrec2022-bib}
\bibliographylanguageresource{languageresource}

\end{document}